\documentclass[11pt]{article}
\usepackage[margin=1in]{geometry}
\usepackage[utf8]{inputenc}
\usepackage[T1]{fontenc}
\usepackage{graphicx}
\usepackage{authblk}
\usepackage[numbers,sort&compress]{natbib}
\usepackage[hidelinks]{hyperref}
\usepackage{microtype}
\title{\textbf{Reasoning in Real World Clinical Care: Why Large Language Models Are Not Yet Safe for Autonomous Clinical Decision Support}}

\author[1,2,3,4]{Shayndhan Sivanathan}
\author[1]{Shravan Nageswaran}
\author[1]{Mehdi Zadem}
\author[1,6]{Ryaan Sultan}
\author[1,7]{Nicolas von Mallinckrodt}
\author[1]{Max Solovyev}
\author[1]{Alexey Matyushkin}
\author[1]{Sumon Sadhu}
\author[8]{Gabriele C DeLuca}
\author[5,9]{Sanjeeva Jeyaretna}
\author[10,11,12]{James Hillis}
\author[13]{Manoj Ramachandran}
\author[14]{Prakash Jayakumar}
\affil[1]{Atman Labs}
\affil[2]{Oxford University Hospitals}
\affil[3]{Nuffield Department of Orthopaedics, Rheumatology and Musculoskeletal Sciences, University of Oxford}
\affil[4]{NIHR Oxford Biomedical Research Centre}
\affil[5]{Department of Neurosurgery, Oxford}
\affil[6]{Imperial College London}
\affil[7]{Technical University of Munich}
\affil[8]{Nuffield Department of Clinical Neurosciences, University of Oxford}
\affil[9]{Nuffield Department of Surgical Sciences, University of Oxford}
\affil[10]{Department of Neurology, Massachusetts General Hospital}
\affil[11]{Mass General Brigham AI, Mass General Brigham}
\affil[12]{Harvard Medical School}
\affil[13]{Barts Health NHS Trust}
\affil[14]{Department of Surgery and Perioperative Care, Dell Medical School, the University of Texas at Austin}

\date{}

\begin{document}
\maketitle

\begin{abstract}
\noindent Large language models (LLMs) now pass medical licensing examinations and, in curated cases, can rival physicians at diagnostic reasoning. These developments have accelerated the use of large language models for symptom assessment and clinical decision support in diagnostic and treatment guidance, administrative documentation, and rules-based alert enhancement. This Perspective concerns the most consequential of these applications: the autonomous triage of self-presenting, undifferentiated patients, with little or no clinician in the loop. For that task, the evidence of safety does not yet exist.

The gap is not in medical knowledge but in the fidelity of clinical evaluation: a model optimized to continue the most probable text is not optimized to act safely when the safe answer is the improbable ``must-not-miss'' diagnosis. Safe triage is not the selection of the most likely diagnosis; it is a sequential decision under asymmetric cost, in which the single catastrophic miss outweighs many false alarms, and the decisive signal may be one the patient has not volunteered---and that the model has not been trained to seek. The core deficit is therefore one of information gathering under uncertainty. Under incomplete histories, LLM systems may fail to show the behaviors safe triage requires: broadening the differential; seeking the missing red flag; lowering the threshold for escalation; deferring judgement until sufficient information is obtained; and escalating concern where high-harm diagnoses remain unexcluded.

These modes of failure for LLMs can be difficult to detect considering that evaluations to date often use complete, well-curated, confidence-gated simulations. These replicas are not suited for testing reasoning under missing information. Further, the application of LLMs under these conditions may be amplified by assistant-like behaviors and positive bias, including credulity, agreeableness, and miscalibration---when these are not constrained by clinical triage logic. We propose the pre-requisite that claims concerning autonomous clinical decision support and evaluation of AI-based reasoning should incorporate incomplete information and asymmetric cost stress testing on validated synthetic and real patient encounters.

\vspace{6pt}
\end{abstract}

\vspace{6pt}
\noindent\textbf{Corresponding author:} Shayndhan Sivanathan MBBS, Oxford University Hospitals, UK. \\ Email: \href{mailto:shay.sivanathan@ouh.nhs.uk}{shay.sivanathan@ouh.nhs.uk}

\section{The Illusion of Readiness: Exam-Passing Is Not Fitness for Practice}

Large Language Models (LLMs) have, by conventional standards, arrived in medicine. They achieve near-perfect scores on medical licensing examinations, and the most capable reasoning models now rival or exceed physicians on demanding diagnostic tasks\cite{ref1}. This performance is established almost entirely on benchmarks: standardized test sets of curated cases against which models are scored. For instance, OpenAI o1 achieved a near-perfect clinical reasoning quality score across validated benchmarks and, in a blinded comparison against attending physicians on real emergency department cases, outperformed them at every diagnostic touchpoint measured\cite{ref1}. Our concern is not clinician-facing aids, but the application now being proposed at the frontline: autonomous triage of undifferentiated patients, with little or no clinician in the loop. Yet two findings expose the distance between this result and fitness for autonomous triage. First, the model's inclusion of ``cannot-miss'' diagnoses was not significantly higher than the physicians it otherwise outscored. Second, even the ``limited information'' at initial triage was structured clinical documentation written by physicians, not the prospective unstructured account of a patient who does not know which details matter. The evaluation was conducted, in the authors' own words, on text-based inputs reliant on the ``careful work of clinicians to curate and `clean up' cases''\cite{ref1}. In a benchmark, that structure is provided by the examiner---the clinician who assembled and completed the case before the model ever saw it. In the clinic, it must be built by asking the right questions, under uncertainty. This is the task current benchmarks rarely set.

The scale of this gap has been measured directly. In a randomized, pre-registered study of 1,298 members of the public, three frontier LLMs (GPT-4o, Llama 3, Command R+) that identified the relevant condition in 94.9\% of scenarios when tested on curated inputs, but in fewer than 34.5\% once real users conveyed their own problem through dialogue\cite{ref2}. Participants assisted by these models fared no better than a control group given no AI at all\cite{ref2}. The knowledge had not changed; the input had. Critically, the failure was not simply that patients describe symptoms poorly. The models often surfaced the correct condition mid-conversation yet failed to elicit the missing detail or ensure it survived into the final decision\cite{ref2}. The model had the knowledge to ask---it did not have the objective to ask. In the clinic, this work is done by the clinician: a structured history is not what the patient volunteers, it is what the doctor builds through directed questioning. When the examiner's role is removed, so is the performance it produced.

This is not a peculiarity of a single study. A systematic review of 445 benchmarks found that the field routinely measures capability through proxies that lack construct validity---tasks that do not represent the phenomenon they claim to assess\cite{ref3}. A licensing score certifies recall on structured questions, not safe reasoning with an unstructured, incompletely disclosing patient. And this failure is not uniform: model robustness degrades most sharply on the rarest and most ambiguous clinical terms\cite{ref4}, concentrating precisely where the atypical, must-not-miss presentations live---the cases that define safety rather than average accuracy. When used to infer fitness for autonomous triage, a licensing score does not merely fall short---it points the wrong way, certifying competence precisely where competence has not been tested.

LLMs are shifting from engineering benchmarks to being proposed at the frontline of care: embedded in search results, offered as symptom-checkers, and piloted as clinician-facing decision support. These settings differ greatly in terms of autonomy, oversight, and risk---a spectrum along which human involvement should be deliberately specified, not assumed\cite{ref5}. Autonomous triage sits at its far end, and it is a step the field is already taking, with `autonomous' primary care and emergency systems now reported\cite{ref6,ref7,ref8,ref9,ref10}.

The deeper problem is what the substrate was never built to do. LLM training optimizes two objectives: pretraining, which teaches the model to continue the most probable text---to extend what is present, but this does not register what is absent; and alignment, which tunes toward human preference and helpfulness. Neither objective represents the logic of safe triage, which is asymmetric: a missed catastrophe and an unnecessary scan are not equivalent errors, and in which the decisive information may be something the patient has not volunteered. Current models remain the most capable substrate available for medical knowledge and synthesis; the deficit is not knowledge. It is that no stage of their training required them to represent the asymmetric cost of being wrong---and no benchmark, with its complete stems and single best answers, test whether they do.

The sections that follow develop this argument: what reasoning from silence requires, why the evidence base has concealed its absence, how assistant behaviors compound it, and what evidence deployment should demand.

\begin{figure}[p]
\centering
\includegraphics[width=\textwidth]{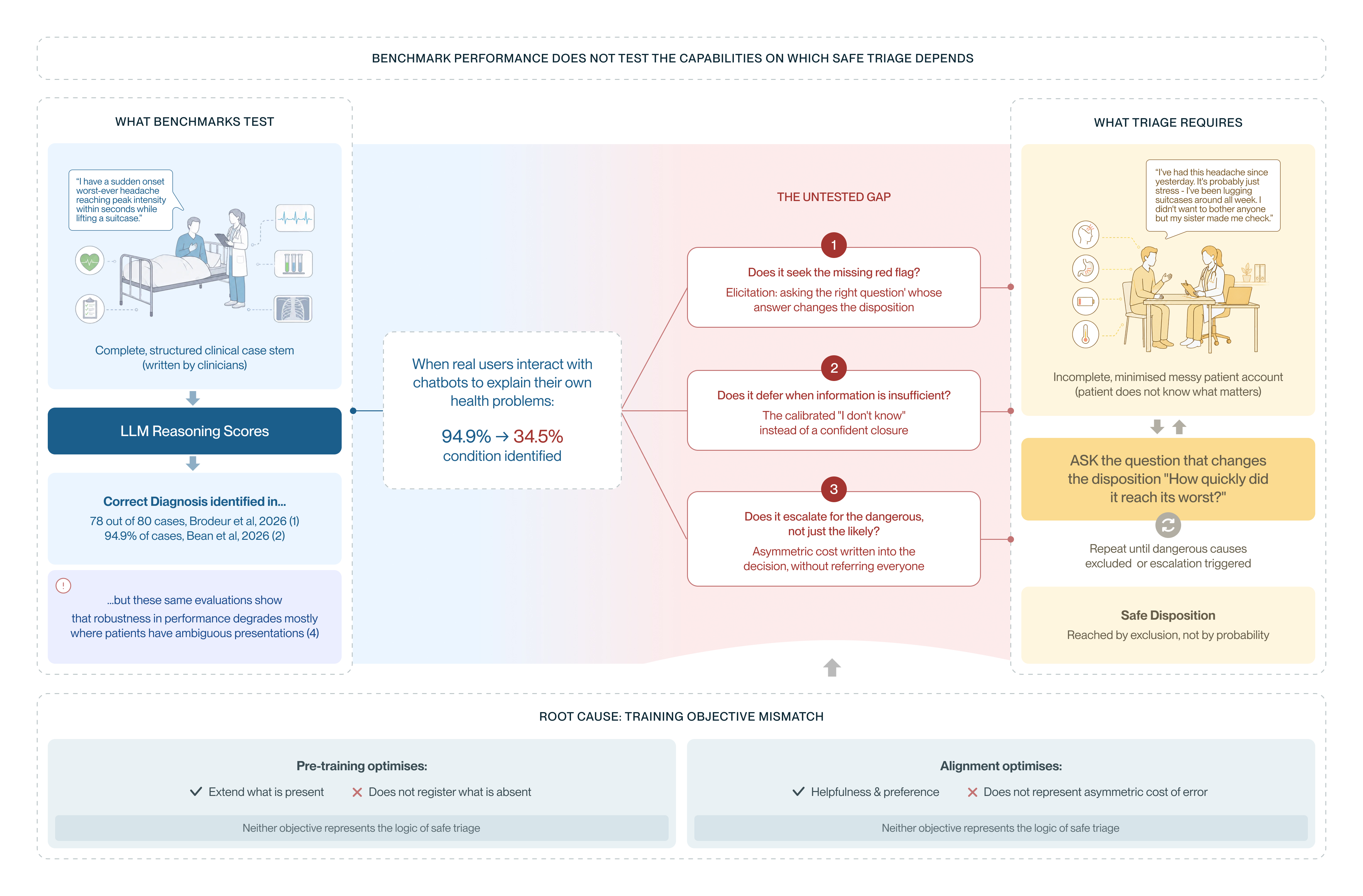}
\caption{Benchmark performance does not test the capabilities on which safe triage depends. Current evaluations (left column) supply a complete, clinician-written case stem and score diagnostic accuracy. Under these conditions frontier models achieve near-perfect reasoning scores and can identify the relevant condition in 94.9\% of cases\cite{ref2}. Real triage (far-right column) begins instead with an incomplete, minimized patient account in which the decisive detail---the red flag that changes the disposition is often not volunteered. When real users conveyed their own problem, condition identification fell from 94.9\% to below 34.5\%\cite{ref2}. This gap represents three behaviors we argue that current standard benchmark rarely scores upon: (1) seeking the missing red flag; (2) deferring when information is insufficient; and (3) escalating on dangerous rather than probable diagnoses. The root cause is an objective mismatch (bottom panel): neither pretraining (next-token continuation) nor alignment (helpfulness and preference) represents the asymmetric cost of a missed diagnosis.}
\label{fig:1}
\end{figure}

\section{Reasoning from Silence: Absence of Evidence Is Not Absence of Disease}

We define reasoning from silence as inference from diagnostically informative missing information: recognizing that what a patient has not said may reflect incomplete elicitation rather than true absence of symptoms.

Consider this representative case: a 41-year-old woman opens a consumer triage tool and reports a headache that started the day before. She is, by temperament, economical: she does not volunteer that the pain reached its peak within seconds, that it was the worst of her life, or that it began suddenly as she lifted a suitcase. The model does not ask. Working from the fragment it is given, it produces a competent looking answer: tension-type headache, migraine and, when prompted, subarachnoid hemorrhage among the possibilities. It then advises that a lumbar puncture is not indicated and assigns her to routine follow-up. Nothing it was told was misinterpreted. The danger lay entirely in what it never asked.

This situation does not reflect a knowledge failure. Given a complete history, the same class of model identifies the correct diagnosis, including subarachnoid hemorrhages among them with 97.5\% accuracy\cite{ref11}. The deficit is not knowledge but objective: a model trained to continue the most probable text has no native representation of the danger it was never told about---and triage demands reasoning about what is absent, not just what is present.

A clinician faced with a sparse history is trained to do three things that a model maximizing for probability does not. They widen the differential rather than narrow it, because uncertainty raises rather than lowers the stakes. They lower the threshold for investigation, because the cost of missing a subarachnoid hemorrhage is not symmetric with the cost of an unnecessary scan---it is larger by orders of magnitude. And where the information needed to decide is simply absent, they defer and ask. In sparse-history evaluations, models have often failed on all three dimensions.

A large stress-test confirms this failure mode at scale. Across 1,000 simulated headache consultations spanning the full spectrum of communication styles and information completeness, frontier models did not grow more cautious as histories grew sparser---they maintained confidence even where incorrect. As information fell from a complete history to 20\% complete information, the rate at which a model recognized that it could not yet decide barely moved, from 5.2\% to 13.8\%, and its differential did not widen at all\cite{ref11}. On a benchmark of whether models recognize the limits of their own knowledge, virtually none could flag a question rendered unanswerable by missing information, responding instead with full confidence\cite{ref12}. The model reasons from what is in front of it; what is absent exerts no force on its confidence.

The pattern is consistent: the model treats what it was not told as though it carries no weight, and what it was told, regardless of how fragmented the information is or who supplies it, as sufficient to make a decision. It conflates absence of evidence with absence of disease. A system fit for triage would do the opposite: treat the unasked question as the most decision-relevant quantity it holds, and the unexcluded catastrophe as the one it is least entitled to ignore. Current systems do neither reliably.

\section{Why the Evidence Cannot See the Failure: The Successes Were Measured on the Wrong Patients}

If this failure is real, why has the evidence not exposed it? Because the studies most often cited as proof of clinical readiness were conducted in the settings least able to do so\cite{ref6,ref7,ref8,ref9,ref10}. The reported successes are real. But they are conditional: measured on populations from which the high-acuity or ambiguous presentations have been excluded.

The pattern begins with the foundational demonstrations. AMIE, the most prominent conversational diagnostic system, matched or exceeded primary-care physicians across most axes of history-taking and diagnostic quality. However, it did so in consultations with validated patient-actors, conducted over a synchronous text interface its own authors call "unfamiliar in clinical practice," within an acknowledged "limited scope of experimental simulated history-taking"\cite{ref13}. Its longitudinal management evaluation repeats the pattern, on cases its authors note "lack much of the chart review to determine clinical history that characterizes real patient care" and a case mix that is "not indicative or representative of a real clinical practice setting"\cite{ref14}. Simulated patients do not withhold what they do not know to mention; an actor reciting a vignette is not the incompletely disclosing patient on whom safe triage turns. The papers state this limitation plainly; the title/headline does not carry it.

Taken into a real clinic, the same system shows what realistic evaluation must remove to proceed safely. In a prospective feasibility study, AMIE conducted pre-visit history-taking for one hundred urgent care patients. However, important eligibility requirements detailed that patients had ``already been determined by clinic triage staff not to need emergency care,'' presented with a single chief complaint, spoke English, and used a computer rather than a phone\cite{ref8}. Any emergencies or ``edge cases'' were triaged out before the system was engaged. Each interaction took place days before the appointment, under continuous supervision by a physician empowered to halt it, and produced not a decision but candidate diagnoses "for the patient to discuss with a provider''\cite{ref8}. This study reflects responsible feasibility design. It is also the exact inverse of autonomous triage: the must-not-miss cases were filtered out at the door, a clinician guarded every consultation, and the system never made the call its results are taken to endorse.

Autonomous evaluation narrows the population by a different route, with the same effect. In the first large-scale, clinician-blinded study of an autonomous primary care system, the AI's leading diagnosis matched the clinician's in 96.3\% of 1,094 encounters\cite{ref6}. But the headline 96.3\% applies only to cases clearing a pre-specified confidence threshold---a filter that, by the authors' account, ``preferentially excluded diagnostically weaker outputs,'' discarding 82.1\% of diagnoses later judged unrelated to the truth, while concordance on the excluded encounters was 87.0\%\cite{ref6}. More revealing is what entered the dataset: encounters ``focused on a single clinical issue'' with a ``sufficiently definite diagnostic impression for routine care, rather than multiple unrelated concerns or a broad, explicitly uncertain differential''\cite{ref6}. The diagnostically ambiguous patient---the one whose safe management depends on reasoning from silence---was defined out of the evaluation before any model was scored. The most autonomous systems narrow it further still: a recent EHR integrated agent evaluated on more than 500 emergency department cases reported outperforming physicians on a matched subset, but the cases were drawn retrospectively from a records database, the patient's history supplied from discharge summaries already written by clinicians, and encounters excluded where that documentation was incomplete\cite{ref7}. Its own authors concede the simulated patient may give ``a more structured account than verbatim, unprompted patient speech in real emergencies.'' The agent was tested, in other words, on histories that had already been taken, cleaned, and completed by the clinicians against whom it was compared.

This is not a catalogue of individual study flaws; it is a structural property of how clinical AI is currently evaluated. Two systematic reviews spanning thousands of studies make the scale of it plain: across 519 LLM evaluations published to early 2024, only 5\% used real patient care data, the rest testing medical knowledge on curated questions such as licensing examinations\cite{ref15}; across 4,609 studies to late 2025, fewer than one in two hundred used real-world data in a prospective randomized design---nineteen trials in all\cite{ref16}. The conditions that constitute triage risk---the rare, the atypical, the undifferentiated, the incompletely disclosed---are systematically thinned from evaluation by acuity screening, confidence gating, actor scripts, supervisory backstops, and case mixes drawn toward everyday illness. An evaluation that removes the catastrophic tail cannot detect a model that mishandles it; a protocol that supplies complete, curated histories cannot reveal a model that fails to ask. The design choices that make these studies tractable are the same choices that render them silent on safety---and silence, here too, is being read as reassurance.

The consequences are already visible where deployment has outpaced evaluation. In a prospective LLM decision-support system study across sixteen Kenyan primary-care clinics\cite{ref9}, the headline results report that LLM-guidance aligned with local guidelines in 99\% of cases. However, actively harmful recommendations still appeared in 7.8\% of encounters---most commonly inappropriate medication recommendations (46.2\%) and omission of critical differential diagnoses (31.6\%)---and clinicians fully or partially adopted the harmful advice in 59\% of those cases, although follow-up review confirmed no resulting adverse patient outcomes\cite{ref9}. High average performance did not prevent tail harm; it obscured it---and the clinicians receiving the recommendations did not reliably catch the failures that remained.

The field has measured what these systems do with the patients it selected for them. It has not yet measured what they do with the patient who walks in undifferentiated, frightened, and only half able to say what is wrong---and that, not licensing exam performance, is the evidence a deployment decision requires.

\begin{figure}[p]
\centering
\includegraphics[width=\textwidth]{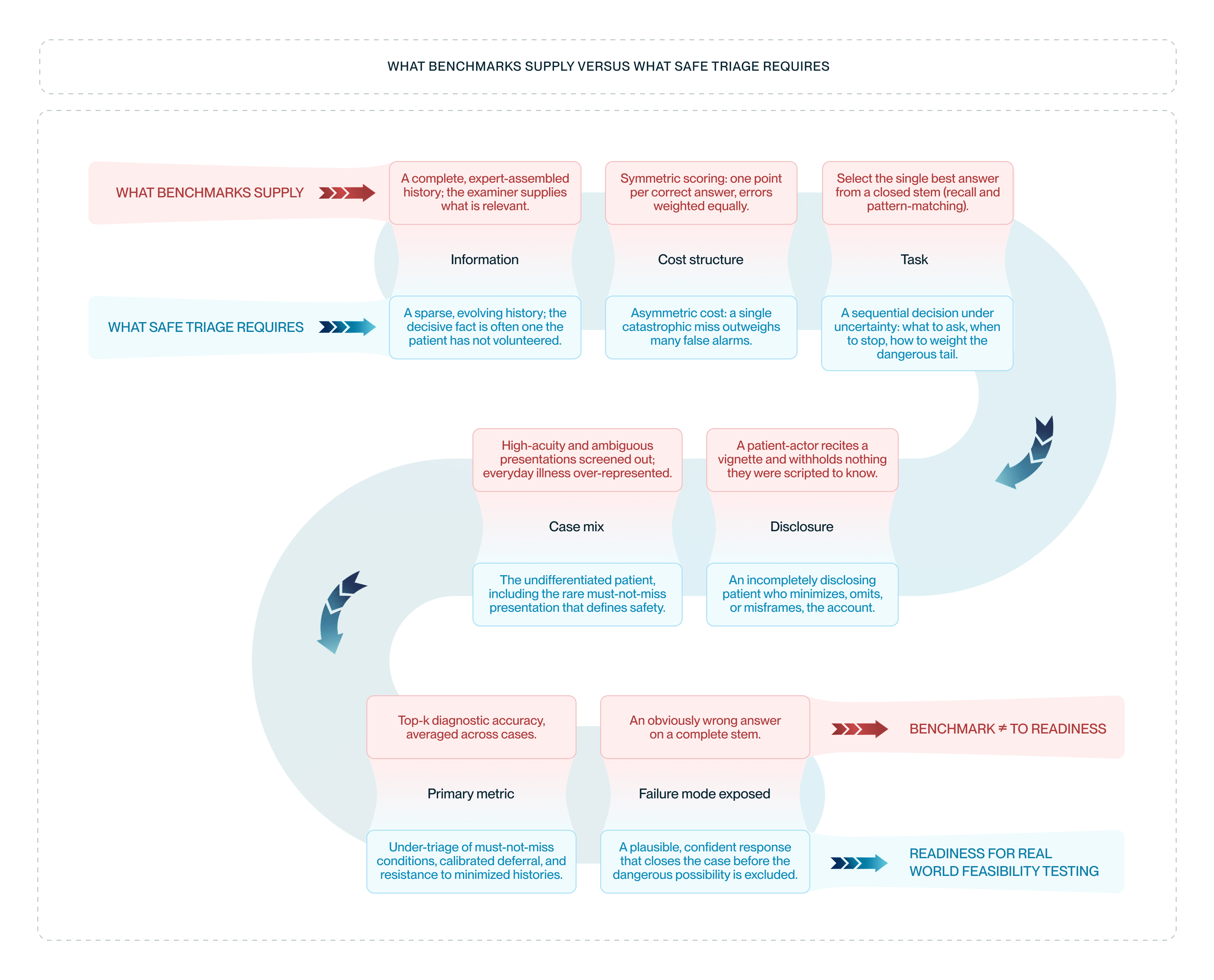}
\caption{What benchmarks supply versus what safe triage requires. Across six dimensions (information; cost structure; task; case mix; disclosure and primary metric), the design choices that make current evaluations tractable are the same choices that render them silent on the behavior that defines triage safety. Benchmarks supply complete, expert-assembled histories, symmetric scoring, closed answer sets, pre-screened case mixes, fully disclosing patient-actors and averaged top-k accuracy. Safe triage turns on sparse, evolving histories, asymmetric error costs, sequential decisions about what to ask and when to stop, the undifferentiated patient, incomplete disclosure and under-triage of must-not-miss conditions. The failure mode each exposes differs accordingly: benchmarks reveal the obviously wrong answer on a complete stem; triage safety is defined by the plausible, confident response that closes the case before the dangerous possibility is excluded. Benchmark success is therefore not evidence of readiness; readiness requires real-world feasibility testing.}
\label{fig:2}
\end{figure}

\section{A Cluster of Safety Inversions: Helpfulness Is Not Safety}

Sections 2 and 3 established the central failure: ``reasoning from silence''---and why the evidence base has not exposed it. That failure does not act alone. It is compounded by a cluster of behaviors rewarded in general-purpose assistants, and these behaviors share a common origin with the core deficit. A model trained to continue probable, preference-aligned text and to condition most heavily on what was said most recently will, by default, accept a premise it is handed, defer to the user's framing, and anchor on its own prior answer. These dispositions are agreeable in an assistant and dangerous in triage. Safe triage requires a system to tolerate doubt, challenge consequential falsehoods, resist premature reassurance and decline to close a case when key information is absent. Four expressions of this single mismatch follow, each surfacing when the patient's account is incomplete, minimized, or wrong.

The first inversion is credulity. When fabricated clinical details: a laboratory value, a physical sign or a diagnosis that does not exist, are embedded in a prompt, models often fail to flag the inconsistency and instead elaborate clinical reasoning around the false premise, doing so in 50\% to 82\% of cases across models and mitigation strategies\cite{ref17}. The vulnerability is reproducible across independent evaluations: on a 10,932 item benchmark of misleading medical context, mean accuracy fell from 71.1\% on clean questions to 38.0\% once a plausible falsehood was introduced\cite{ref18}. Susceptibility is greatest when the false claim is delivered in authoritative clinical prose rather than flagged as opinion or dressed in obvious rhetorical fallacy\cite{ref17,ref18}. A safe history-taker treats an implausible or consequential claim as something to probe. A fluent assistant may treat it as material to accept and extend.

The second inversion is agreeableness. Tuning a model to be warmer and more empathetic may improve user experience, but in consequential tasks it can increase error rates and the likelihood of affirming a user's incorrect belief, especially when the user expresses distress\cite{ref19}. This matters because triage often begins with a patient's interpretation: ``it is probably just stress,'' ``I do not want to waste anyone's time,'' or ``I am sure it is not serious.'' Safe triage requires the system to resist reassurance when a high-harm diagnosis remains unexcluded. A model optimized for agreeable interaction may instead align with the patient's minimization.

The third inversion is miscalibration under uncertainty. The relevant confidence signal in triage is not confidence in the most likely diagnosis, but confidence that dangerous alternatives have been sufficiently excluded. LLMs can show choice supportive bias after committing to an answer and over-correction when challenged \cite{ref20}. In direct tests of judgment revision as new information arrives, leading models fell short of attending physicians, were systematically overconfident, and rarely selected a neutral "this does not change my assessment" response\cite{ref21}. A system whose expressed certainty is not calibrated to missingness, new evidence and clinical stakes cannot safely use confidence as a disposition signal.

These behaviors converge on omission. On a specialist-validated benchmark of 4,249 real management decisions, the potential for severe harm reached 22.2\% of cases, and 76.6\% of errors were omissions: failures to recommend an action that was needed rather than recommendations of an action that was not\cite{ref22}. Whether reasoning from silence directly or the downstream consequence of the preceding inversions, the failure is the same: the system need not hallucinate a dangerous intervention to harm a patient; it can fail by accepting the incomplete premise, affirming minimization, expressing unwarranted certainty, or omitting the action that would keep a must-not-miss diagnosis in play.

Across credulity, agreeableness, miscalibration and omission, the pattern is not that helpfulness is always unsafe. It is that assistant-like behaviors become unsafe when they are not constrained by the clinical logic of triage. The remedy is not a more knowledgeable model; it is a different discipline---one that refuses to close the case until missing high-harm variables have been sought, excluded, or escalated. What that discipline requires of evaluation is the subject of the next section.

\section{What Would Actually Demonstrate Fitness: From Accuracy Scores to Safety Stress-Tests}

If the deficit were one of knowledge, the remedy would be a better model. It is not. The preceding sections locate the failure at two levels that no amount of scale addresses on its own: the objective an LLM optimizes, and the evaluation by which we decide it is ready. A model trained to continue the most probable text is not, by default, optimizing for safe action under asymmetric cost; and an evaluation built on curated, complete, pre-triaged cases cannot tell us whether it does. Both must be confronted; the second one is actionable now, and is the precondition for trusting any claim about the first.

Nor is the remedy simply to make the model talk more. Prompting a model to gather additional history can lower diagnostic accuracy by as much as 11.3\%\cite{ref23}; asking is not the same as asking well. What distinguishes the two is structure: in a randomized study of 13,917 people, agents conducting a systematic symptom interview before committing to a diagnosis outperformed default user-guided exchanges, producing differentials more accurate than clinicians working from the same dialogue\cite{ref24}. Elicitation, done deliberately, is the difference---but it is a deliberate policy layered onto the model, not a property the training objective supplies. And accuracy on everyday illness is not yet safety on the catastrophic tail: a safe questioner must choose the question whose answer would most change what happens next. This is a problem of the value of information, and it is genuinely hard: a system that asks the single most informative question at each step can still be provably suboptimal when no individual question is decisive on its own\cite{ref25}. The competence triage demands are not better recall but a different kind of computation---sequential, selective, and weighted toward the catastrophic tail.

What follows is a set of requirements for evaluation that current practice does not meet. An adequate evaluation must do three things.

First, it must withhold information by design. Where existing frameworks supply the history and ask the model to reason over it---even the most advanced dynamic simulators still furnish the clinical facts\cite{ref26}---a fit-for-purpose evaluation must generate patients whose histories are deliberately incomplete, and vary that incompleteness systematically, with the ground-truth diagnosis and correct disposition known to the evaluator but not the model. Only by controlling what the patient does not say can one measure whether a system reasons from silence or merely completes the probable. Adaptation-based proposals leave untouched: where such work asks whether a model can be made to fit the clinical context it is given\cite{ref23}, the prior question is whether it reasons safely about the danger the context omits. Supplying better context changes what the model is told, not how it weights what it was not.

Second, it must score the right objective. Top-k diagnostic accuracy, the field's default metric, is silent on safety: a model can carry the correct diagnosis in its differential and still refuse the investigation that would confirm it. A safe-triage evaluation must instead reward the behaviors the preceding sections showed to be missing: widening under uncertainty, lowering the threshold for escalation, deferring when information is insufficient, and resisting minimized or adversarial histories. Under-triage of must-not-miss conditions, the calibrated "I don't know," turns-to-safe-disposition, and resistance to the inversions described in Section 4 belong in the scorecard alongside accuracy---and the asymmetric cost of error must be written into the metric itself, not averaged away. The counterweight belongs in the same scorecard: over-triage and unnecessary escalation carry real costs---crowded emergency departments, cascades of investigation, alarm fatigue---and a system that responds to uncertainty by referring everyone has not learned to reason from silence; it has replaced one indiscriminate rule with another. Because a model's own account of its reasoning cannot be trusted to reveal these properties\cite{ref27}, the evaluation must be behavioral: scoring what the system does, not what it says it did.

Third, it must validate a simulator before trusting it. An evaluation built on synthetic patients inherits the blind spots of whatever generated them; the atypical presentation no model imagines is both absent from the test and lethal in the clinic. Simulator realism therefore cannot be assumed --- it must be demonstrated, in three escalating layers: face validity, ideally via a blinded test in which clinicians cannot distinguish synthetic transcripts from real ones; distributional validity, confirming that the rate and variety of rare, must-not-miss presentations match real epidemiology rather than the generator's expectations; and predictive validity, evidence that performance on the simulator forecasts performance on real patients with real outcomes. Until distributional and predictive validity are established, strong simulator performance is a screening hurdle a system must clear, not a warrant for deployment.

None of this requires waiting for a new architecture. It requires deciding that the evidence that would justify autonomous triage is evidence of safe behavior under realistic uncertainty, not performance on curated cases that could not expose the failure in the first place. This position is no longer fringe: a multi-stakeholder expert panel recently concluded that prevailing standards largely monitor safety and process compliance but "do not address effectiveness"---the demonstration of improved outcomes --- and that retrospective accuracy "is not necessarily indicative of how care decisions will be influenced by the tool in practice"\cite{ref28}, while this journal's editors have put it more plainly still: the stronger the claim, the stronger the evidence must be\cite{ref29}. The claim that an LLM can be trusted to triage the undifferentiated patient is among the strongest in medicine. The evidence remains mismatched to the strength of the claim. Closing that gap is the precondition for everything that follows.

\begin{figure}[p]
\centering
\includegraphics[width=\textwidth]{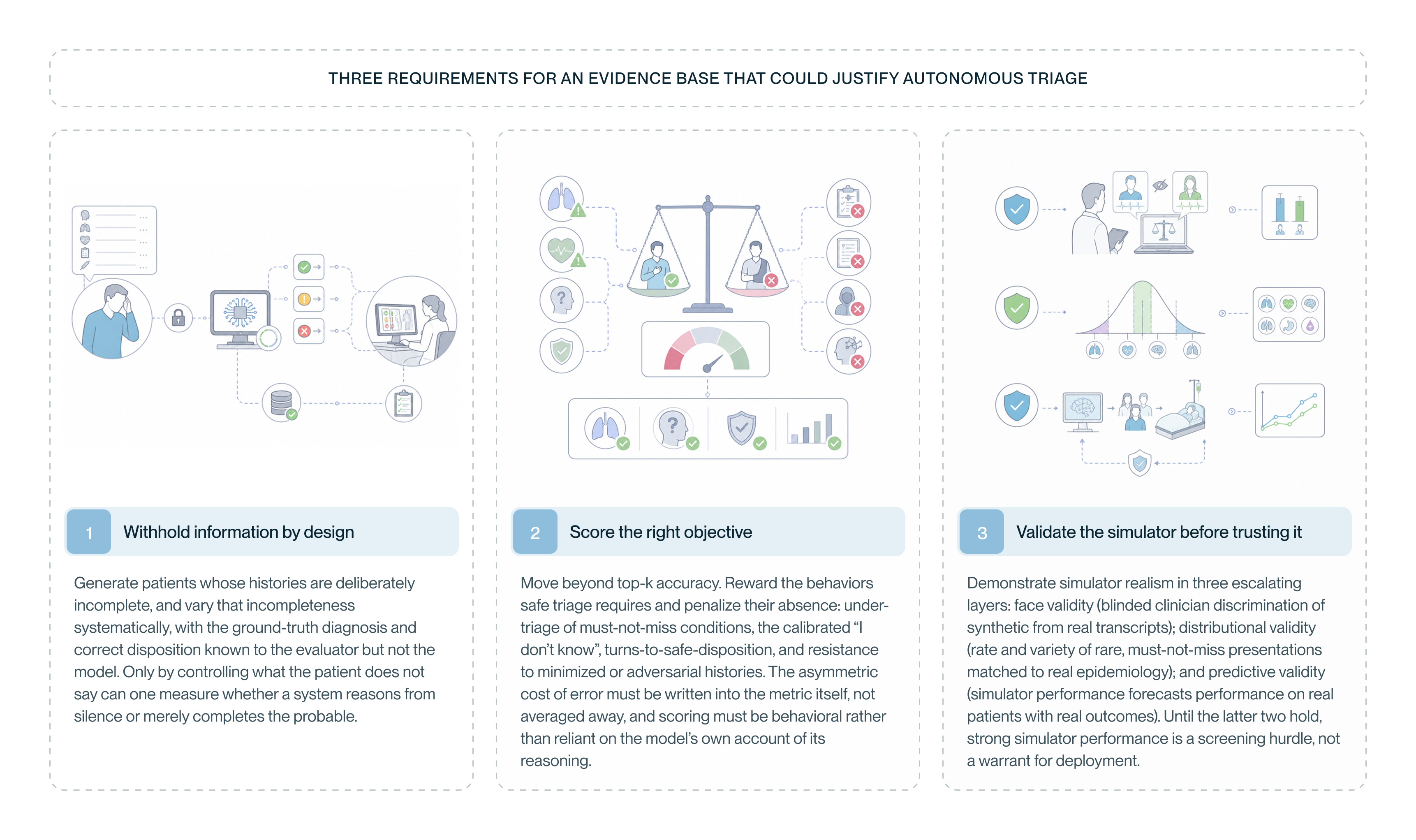}
\caption{Three requirements for an evidence base that could justify autonomous triage. \cite{ref1} Withhold information by design: generate synthetic patients whose histories are deliberately and systematically incomplete, with the ground-truth diagnosis and disposition known to the evaluator but not the model---only by controlling what the patient does not say can an evaluation measure whether a system reasons from silence. \cite{ref2} Score the right objective: move beyond top-k accuracy to reward the behaviors safe triage requires, penalizing under-triage of must-not-miss conditions and unnecessary escalation alike, with the asymmetric cost of error written into the metric and scoring based on what the system does rather than its own account of its reasoning. \cite{ref3} Validate the simulator before trusting it, in three escalating layers: face validity (blinded clinician discrimination of synthetic from real transcripts), distributional validity (rare, must-not-miss presentations matched to real epidemiology) and predictive validity (simulator performance forecasts performance on real patients with real outcomes). Until the latter two hold, strong simulator performance is a screening hurdle, not a warrant for deployment.}
\label{fig:3}
\end{figure}

\section*{Conclusion}

The case for autonomous LLM triage currently rests on evidence that does not measure the task most relevant to safety. Examination performance and curated vignette accuracy demonstrate medical knowledge and fluent reasoning over complete inputs; they do not show whether a system can manage danger under incomplete and ambiguous information, where the decisive fact may be one the patient has not volunteered. The available evidence raises concern that performance can degrade when histories become sparse: models may include a must-not-miss diagnosis in the differential while failing to seek the missing red flag, recommend the indicated investigation, or preserve an urgent disposition.

This situation does not simply reflect a shortfall of medical knowledge that scale alone will close. It is an evidentiary gap around a specific behavior: safe, sequential, harm-weighted action when information is missing. The characteristic risk is therefore not only an obviously false answer, but a plausible and confident response that closes the case before the dangerous possibility has been excluded.

The remedy is to change what is accepted as evidence. Before any system is trusted to triage the undifferentiated patient without a clinician in the loop, it should demonstrate safe behavior under realistic incomplete-information conditions: hidden variables; minimization; adversarial or third-party histories; harm-weighted scoring. For autonomous triage, that evidence is not benchmark accuracy alone; it is demonstrated safety when the patient has not yet said the thing that matters.

\section*{Declarations}

\paragraph{Ethical approval}

This review did not require ethical approval.

\paragraph{Consent for publication}

All authors consent for the following manuscript to be peer-reviewed and considered for publication

\paragraph{Data Availability Statement}

The data set used for this review will be shared upon request from the study authors.

\paragraph{Conflict of interest} All authors declare no conflict of interest.

\paragraph{Funding Statement}

This research did not receive any specific grant from funding agencies in the public, commercial or not-for-profit sectors. The research was supported by the National Institute for Health Research (NIHR) Oxford Biomedical Research Centre (BRC). The views expressed are those of the author(s) and not necessarily those of the NHS, the NIHR or the Department of Health.

\paragraph{Author contributions}

SS, SN, MZ, RS, NVM, MS, AM, SS, PJ and conceived and designed the article. SS, SN, MZ, RS and SS wrote the main manuscript text. PJ amd MR supervised the project. AM made the  figures and contributed to writing sections of the manuscript. All authors critically revised the manuscript for intellectual content. All authors reviewed and approved the final manuscript.

\paragraph{Disclosures}

The authors have no personal, financial, or institutional interest in any of the materials described in this article.

\end{document}